\documentclass[pdflatex,sn-mathphys-num]{sn-jnl}

\usepackage{graphicx}%
\usepackage{multirow}%
\usepackage{amsmath,amssymb,amsfonts}%
\usepackage{amsthm}%
\usepackage{mathrsfs}%
\usepackage[title]{appendix}%
\usepackage{xcolor}%
\usepackage{textcomp}%
\usepackage{manyfoot}%
\usepackage{booktabs}%
\usepackage{algorithm}%
\usepackage{algorithmicx}%
\usepackage{algpseudocode}%
\usepackage{listings}%
\usepackage{array}%
\usepackage{tabularx}      
\usepackage{makecell}
\usepackage{ulem}         
\usepackage{subcaption}   
\usepackage{geometry}     

\theoremstyle{thmstyleone}%
\theoremstyle{thmstyletwo}%

\theoremstyle{thmstylethree}%

\begin{document}

\title[GBDP]{MoDE-Boost: Boosting Shared Mobility Demand with Edge-Ready Prediction Models\footnote{A preliminary version of this paper appears in \cite{TziorvasSharedMob}}}


\author*{\fnm{Antonios} \sur{Tziorvas}}\email{atzio@unipi.gr}

\author{\fnm{George S.} \sur{Theodoropoulos}}\email{gstheo@unipi.gr}

\author{\fnm{Yannis} \sur{Theodoridis}}\email{ytheod@unipi.gr}

\affil{\orgdiv{Department of Informatics}, \orgname{University of Piraeus}, \orgaddress{\city{Piraeus}, \country{Greece}}}




\abstract{Urban demand forecasting plays a critical role in optimizing routing, dispatching, and congestion management within Intelligent Transportation Systems. By leveraging data fusion and analytics techniques, traffic demand forecasting serves as a key intermediate measure for identifying emerging spatial and temporal demand patterns. In this paper, we tackle this challenge by proposing two gradient boosting model variations, one for classification and one for regression, both capable of generating demand forecasts at various temporal horizons, from 5 minutes up to one hour. Our overall approach effectively integrates temporal and contextual features, enabling accurate predictions that are essential for improving the efficiency of shared (micro-)mobility services. To evaluate its effectiveness, we utilize open shared mobility data derived from e-scooter and e-bike networks in five  metropolitan areas. These real-world datasets allow us to compare our approach with state-of-the-art methods as well as a Generative AI-based model, demonstrating its effectiveness in capturing the complexities of modern urban mobility. Ultimately, our methodology offers novel insights on urban micro-mobility management, helping to tackle the challenges arising from rapid urbanization and thus, contributing to more sustainable, efficient, and livable cities.}




\maketitle
\clearpage
\section{Introduction}\label{sec:intro}

Urban shared mobility, often framed as Mobility‑as‑a‑Service (MaaS) \cite{MLADENOVIC202112}, integrates heterogeneous transportation options, including public transit, micro‑mobility services (e.g. bike‑ and scooter‑sharing), and commute‑based schemes, such as car‑pooling. The rapid proliferation of real‑time data analytics within Intelligent Transportation Systems (ITS) has heightened the demand for accurate mobility‑pattern forecasting, which can alleviate congestion, reduce travel times, and improve road safety in increasingly complex urban environments.

Particularly, in shared‑mobility contexts, encompassing ride‑hailing, bike- and scooter‑sharing, spatiotemporal demand prediction is essential for efficient resource allocation, minimized passenger waiting times, and optimal fleet deployment. Reliable forecasts also underpin smart‑city initiatives by informing human‑centric infrastructure design and supporting sustainable development through data‑driven decisions in public‑transit planning \cite{Caggiani2018AMF, GNNPTDP2020}, and emergency‑service provisioning \cite{Vemuri2024EnhancingPT}.

Shared Micro-mobility Demand Forecasting (SMDF) specifically addresses the problem of predicting when and where users will request related assets (e‑scooters, bicycles) in urban areas. Effective SMDF is crucial for operators to manage and rebalance fleets, as well as for municipal planners to alleviate congestion. The task involves extracting latent mobility patterns from high‑frequency spatio‑temporal data and translating them into actionable forecasts that guide resource allocation. Robust SMDF solutions can therefore improve service availability, enhance user experience, and contribute to more efficient, sustainable urban transportation systems.

\begin{figure}[htb]
    \centering
    \includegraphics[width=\textwidth]{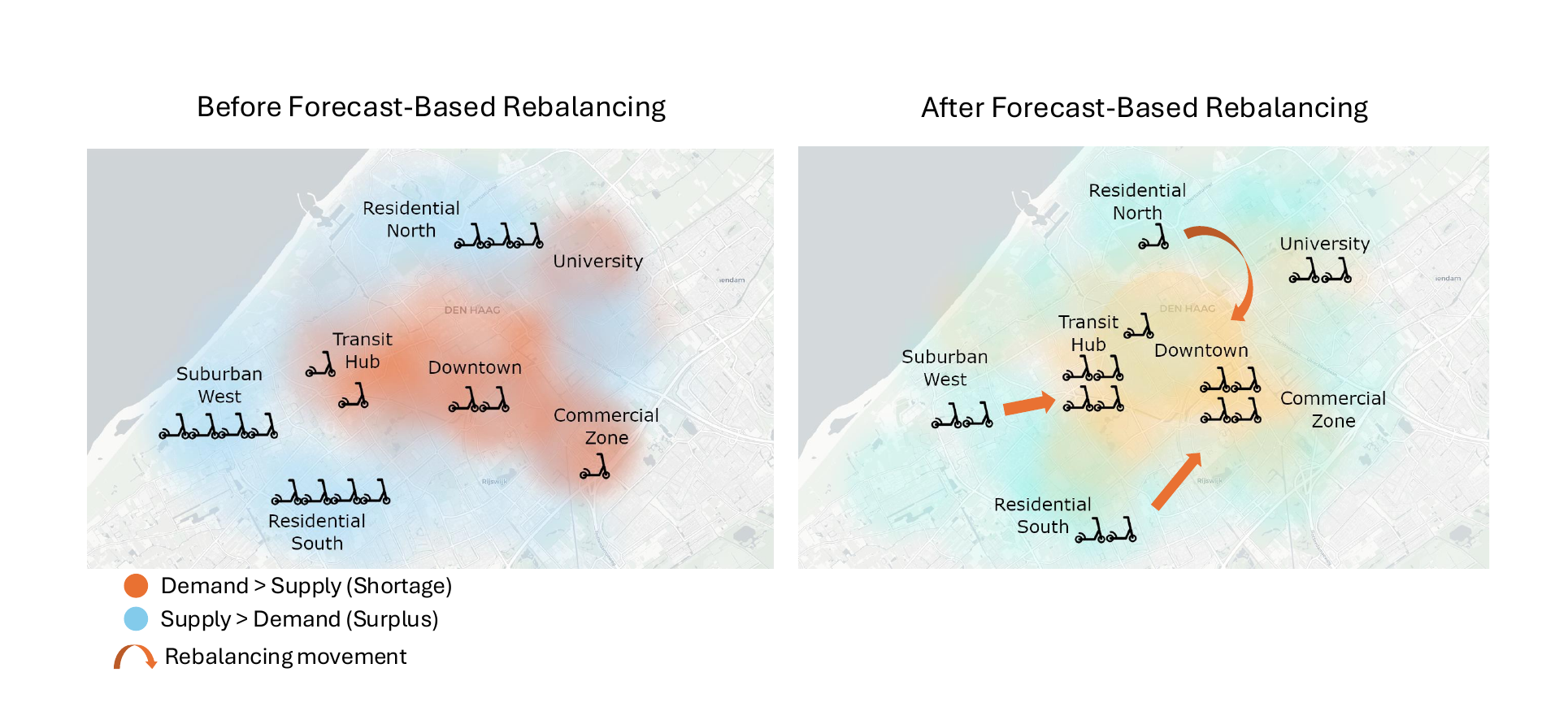}
    \caption{Impact of forecast‑driven fleet rebalancing in Rotterdam. Left panel: pre‑rebalancing fleet with red circles (demand $>$ supply) and blue circles (demand $<$ supply) indicating shortages and surpluses, respectively. Right panel: post‑rebalancing after moving scooters (orange curved arrows) – shortage areas turn light orange, surplus circle fade to teal}
    \label{fig:shortage-supply}
\end{figure}

Figure \ref{fig:shortage-supply} visualizes the consequences of inaccurate demand forecasting for the e‑scooter fleet in Rotterdam and the gains achieved through a forecast‑driven rebalancing operation. The left panel depicts the pre‑rebalancing situation: red, semi‑transparent circles highlight three high‑demand zones (the city center/Downtown area, the Transit Hub, and the Commercial Zone) where the number of requested rides exceeds the available scooters, while light blue circles indicate the two surplus areas (the suburban/residential areas) in which many vehicles remain idle. On the other hand, the right panel shows the post‑rebalancing configuration after moving scooters from the surplus districts to the shortage hot-spots. Curved orange arrows denote the transferred scooters, the shortage overlays become fainter orange gradients (signaling that demand is largely satisfied), and the surplus zones shrink to a teal shade, signaling a more appropriate surplus.

A rich body of literature has explored Machine Learning (ML) approaches for detecting and forecasting spatio‑temporal patterns in time‑series data. Notable examples include Long Short‑Term Memory (LSTM) networks \cite{Hu2021ASL}, Graph Neural Networks (GNNs) \cite{Yu2017SpatiotemporalGC, Zheng2019GMANAG}, and Diffusion‑based models \cite{Yang2024ASO, Li2017DiffusionCR}. These methods commonly employ a Spatio‑Temporal Graph (STG) representation, where nodes correspond to spatial locations and edges encode temporal relationships. Advanced architectures such as Spatio‑Temporal Graph Neural Networks (STGNNs) \cite{Liao2022TaxiDF} and Spatio‑Temporal Graph Convolutional Networks (STGCNs) \cite{Geng2019STMGCN} have demonstrated strong predictive capabilities. Despite their accuracy, such models often suffer from high computational complexity because they must jointly capture intricate spatial correlations and temporal dependencies \cite{Pelekis2014-fw, Wu2019ACS}. This computational burden hampers real‑world deployment, especially for large‑scale, city‑wide forecasting tasks. While model‑compression techniques (e.g., pruning, quantization) can reduce inference latency, they typically have unwanted side effects \cite{Hooker2020Bias} or result in a trade‑off between efficiency and predictive performance \cite{LibanoQuant2020}.

To address this gap and focusing on the SMDF problem framed earlier, in this paper, we propose a gradient‑boosted‑tree (GBT) framework that leverages the proven efficacy of tree‑based ensembles on structured, tabular data \cite{Grinsztajn22}. Our approach consists of two key components: (i) a robust feature‑extraction pipeline that encodes both temporal dependencies and contextual features into a compact tabular representation, and (ii) a GBT model capable of addressing the problem at hand via either classification (e.g., demand levels “Low”, “Medium”, “High”) or regression (i.e., absolute demand values). The resulting framework delivers competitive forecasting accuracy while maintaining low computational overhead, rendering it suitable for large‑scale, real‑time deployment on resource-constrained environments as well. We evaluate the proposed methodology on real‑world micro‑mobility datasets from five metropolitan areas, two from USA (namely, NY and Chicago) and three from The Netherlands (namely, Amsterdam, The Hague, and Rotterdam). Our experimental results indicate that the proposed GBT‑based models adapt effectively to the unique characteristics of each urban area, achieving performance comparable to, or even surpassing, state‑of‑the‑art STGNNs with a fraction of the computational cost. In summary, the contributions of the paper are as follows:

\begin{itemize}
    \item We formally define the SMDF problem, in two variations, namely point- and region-based, according to the desired aggregation level. 
    \item We present a robust feature-extraction pipeline that captures both inter-region relationships and intra-region mobility behaviors, in a compact tabular representation.
    \item We propose a novel GBT framework, called Mobility Demand Edge-ready Boosting Predictor (MoDE-Boost), for spatio-temporal demand forecasting in micro-mobility datasets, in two variations: classification- and regression-based (hence, C-MoDE-Boost and R-MoDE-Boost, respectively).
\end{itemize}

The remainder of this paper is organized as follows. 
Section \ref{sec:relatedwork} provides a concise survey of the most relevant literature. 
Section \ref{sec:proposedapproach} details our forecasting methodology, describing the preprocessing pipeline, feature‑engineering strategies, and the full configuration of the MoDE-Boost architecture. 
Section \ref{sec:experimentalstudy} presents our experimental study, targeting two different layers of the computing continuum: processing at the cloud and at the edge, respectively (with the latter focusing mostly on latency and memory footprint). 
Finally, Section \ref{sec:conclusions} summarizes the main contributions, discusses the broader implications of our findings, and outlines promising directions for future research.  
\section{Related Work}\label{sec:relatedwork}

A wide variety of modeling approaches have been proposed to tackle the challenges related to spatio-temporal forecasting.
In this section, we identify and review such approaches in the context of shared mobility demand forecasting by categorizing them into three categories: (i) classical statistical and ML methods, (ii) deep‑learning models that exploit convolutional architectures, and (iii) GNN‑based frameworks. This taxonomy highlights the evolution from feature‑engineered baselines toward representations that explicitly model spatial heterogeneity and temporal dynamics.

\textbf{Statistical and ML Approaches}: Early investigations relied on classical regressors and tree‑based ensembles, occasionally integrating them with shallow neural components. Lee et al. \cite{Lee2024} compared Multiple Linear Regression, Random Forest regression, and an MLP for bike‑sharing demand prediction. Their workflow emphasized rigorous preprocessing, including the removal of low‑correlation predictors and the incorporation of lagged weather and usage variables across hourly and weekly windows. Sohrabi and Ermagun \cite{Sohrabi2021DynamicBS} introduced a two‑stage pattern‑detection pipeline in which “target” and “candidate” traffic profiles are constructed from 15‑minute aggregates; similarity is then assessed with a weighted Euclidean K‑nearest‑neighbor metric that encodes historical, temporal, and spatial cues. Caggiani et al \cite{STClusterFFBSS} introduced a spatio‑temporal framework that first aggregates bike‑usage patterns via wavelet‑based hierarchical clustering followed by k‑means spatial clustering, and then forecasts cluster‑level demand with a Non‑Linear Autoregressive Neural Network (NARNN), enabling an approximate vehicle‑routing formulation for bike relocation.These works demonstrate that statistical baselines, when enriched with carefully engineered features, remain competitive baselines for short‑term demand prediction.

\textbf{Deep learning models with convolutional architectures}: Convolutional Neural Networks (CNNs) and their variants have become the de‑facto choice for learning hierarchical spatial representations by treating urban spaces as image‑like grids. Zhang et al. \cite{ZhangSTResNet} presented the Deep Spatio‑Temporal Residual Network (ST‑ResNet), which partitions a city into an $I \times J$ grid and models inflow/outflow as a two‑channel tensor. Separate residual streams capture closeness, daily period, and weekly trend, and their outputs are fused by a parametric‑matrix‑based weighting scheme before concatenation with external factors (e.g., weather, events). Yao et al. \cite{Yao2019STDN} introduced the Spatial‑Temporal Dynamic Network (STDN), where a Local Spatial‑Temporal Network (LSTN) employs local CNNs to encode spatial interactions and LSTMs for short‑term temporal dependencies. A Flow‑Gating Mechanism modulates the CNN kernels with a sigmoid‑gated flow signal, while a Periodically Shifted Attention Mechanism addresses temporal shifts across daily/weekly cycles. Li et al. \cite{Li2022STMN} proposed the Spatial‑Temporal Memory Network (STMN), comprising three independent Conv‑LSTM modules that learn closeness, period, and trend representations; a feature‑fusion module (STMN‑WCAT) that performs weighted concatenation proved most effective. Yang and Li \cite{YangLiProphetBiLSTM} proposed a Prophet‑BiLSTM hybrid in which Prophet captures non‑linear seasonalities (daily, weekly, holidays) and external factors (weather, wind), while a bidirectional LSTM learns bidirectional temporal dependencies; the two forecasts are fused via least‑squares‑derived weights. Collectively, these studies illustrate how convolution‑based designs can jointly capture local spatial context and multi‑scale temporal patterns.

\textbf{GNN‑based Approaches}: Graph-structured models provide a natural representation of spatial relationships within a network of stations. Feng and Liu \cite{FengLiuASTN} designed an Adaptive Spatial‑Temporal Network (ASTN) that first partitions stations into patches and applies a Group Multi‑head Self‑Attention (G‑MSA) mechanism to learn both local and global dependencies. Two adaptive adjacency matrices, one derived from historical rides, the other from exogenous factors, are fused via graph convolutional layers. Lee et al. \cite{Lee2019TGNet} introduced the Temporal‑Guided Network (TGNet), which employs permutation‑invariant graph operations to aggregate neighboring region features, while a Temporal‑Guided Embedding encodes explicit time‑of‑day, day‑of‑week, and holiday cues that are concatenated with demand inputs. Geng et al. \cite{Geng2019STMGCN} presented a Spatio‑Temporal Multi‑Graph Convolution Network (ST‑MGCN) that simultaneously models three relational graphs (spatial proximity, POI similarity, and transportation connectivity). Temporal dynamics are captured by a Contextual Gated Recurrent Neural Network (CGRNN) that fuses historical demand with graph‑derived contextual embeddings and applies attention‑based re‑weighting before a shared RNN decoder. These GNN‑based frameworks consistently outperform purely convolutional baselines, especially in settings where long‑range, irregular dependencies dominate the mobility patterns.

Although the aforementioned works constitute the state of the art for shared micro-mobility systems, they predominantly focus on point-based demand (i.e., at station level) and do not consider region‑based forecasting, where clusters of stations may be aggregated over a predefined spatial extent. Consequently, a direct comparison with our methodology is limited to the point-based variation of the SMDF problem, and for this purpose, we select ASTN \cite{FengLiuASTN} as the current state-of-the-art method. On the other hand, for the region-based variation of the problem at hand, we compare our proposed framework with a Generative AI-based approach based on TimeGPT \cite{garza2023timegpt1}.
\section{MoDE-Boost: a Gradient Boosting-based, Edge-ready Framework for Shared Micro-mobility Demand Forecasting}\label{sec:proposedapproach}

In this section, we define the shared mobility demand forecasting problem (Section \ref{sec:problemdef}) and then, we proceed with a detailed presentation of the so-called \textbf{Mo}bility \textbf{D}emand \textbf{E}dge-ready \textbf{Boost}ing Predictor (MoDE-Boost), our approach for addressing the problem at hand (Section \ref{sec:methodology} and Section \ref{sec:feature-extraction} provide an overview of the general architecture and a detailed presentation of the features participating in the MoDE-Boost model, respectively). 

\subsection{Problem Definition}\label{sec:problemdef}
The goal of this work is to forecast the demand for shared micro-mobility services (e‑scooters, e‑bikes) at a spatial unit of interest—could it be an individual dock station or an aggregated administrative region—over a future time horizon, aka the SMDF problem framed in Section \ref{sec:intro}. In particular, we support two types of spatial resolution: point- and region- based, hence the \textbf{p}-SMDF and the \textbf{r}-SMDF problem, respectively.
\paragraph{Prerequisites}
Formally, let $\mathcal{V}$ denote a set of spatial entities (points or non-overlapping regions), $D_t^{A}\in\mathbb{N}$ the non‑negative demand value of micro-mobility vehicles within (at) region (point, respectively) $A$ at time $t$, and $\mathbf{X}^{A}_{1:t}$ the \textit{historical demand vector}
\[
    \mathbf{X}^{A}_{1:t}
    =\bigl(D^{A}_{1},\,D^{A}_{2},\,\dots ,\,D^{A}_{t}\bigr)^{\!\top}
    \in\mathbb{N}^{t}.
\]
associated with a specific target spatial entity\footnote{For sake of consistency, in the discussion that follows we use the (unified) term "spatial entity" to refer to either regions (for the r-SMDF problem) or points (for the p-SMDF problem, respectively).} $A$, along the discrete time axis $\mathcal{T}=1,\dots,t$
\paragraph{Forecasting task}
Given the historical demand vector $\mathbf{X}_{1:t}^{A}$ for all $A\in\mathcal{V}$, the Shared Micro-mobility Demand Forecasting problem (r‑SMDF for the region‑based case, p‑SMDF for the point‑based case) is to predict the demand value $D_{t+H}^A$, $H$ steps ahead:
\[
    \hat D_{t+H}^{A}=F\bigl(\mathbf{X}_{1:t}^{A}\bigr),\qquad
    A\in\mathcal{V},
\]
or, equivalently,
\[
    \hat{\mathbf{X}}_{t+H}
    =F\bigl(\mathbf{X}_{1:t}\bigr).
\]
where $\mathbf{X}_{1:t}=\bigl[\mathbf{X}^{A}_{1:t}\bigr]_{A\in\mathcal{V}}$ aggregates the historical vectors of all entities.

\begin{figure}
    \centering
    \includegraphics[width=\linewidth]{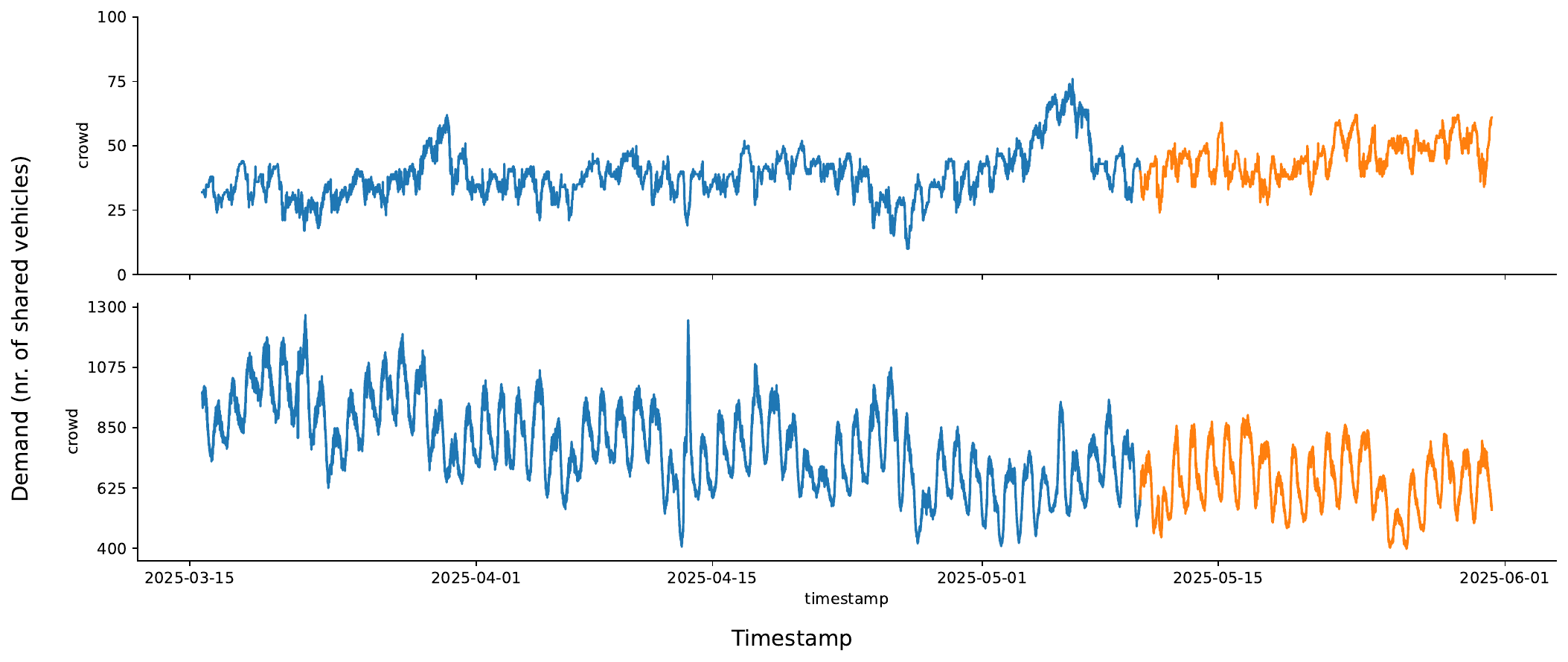}
    \caption{Two indicative timeseries for two regions of Rotterdam: Overschie (top) and Rotterdam Centrum (bottom). The x-axis denotes time, while the y-axis indicates the measured demand values in number of shared vehicles.}
    \label{fig:compare}
\end{figure}

To ground the above definition in a real-world scenario, we illustrate an example derived from one of the datasets used in our experimental study, micro-mobility demand in Rotterdam, the Netherlands (to be detailed in Section \ref{sec:experimentalstudy}). As shown in Figure \ref{fig:compare}, the SMDF problem involves predicting the future values (orange) of each city district’s mobility demand timeseries, given its current and historical observations (blue), while incorporating relevant temporal and contextual features. Accurate forecasting requires uncovering latent patterns—such as seasonality and recurring trends—embedded within these timeseries.

\begin{figure}[!ht]
    \centering
    \includegraphics[width=\columnwidth]{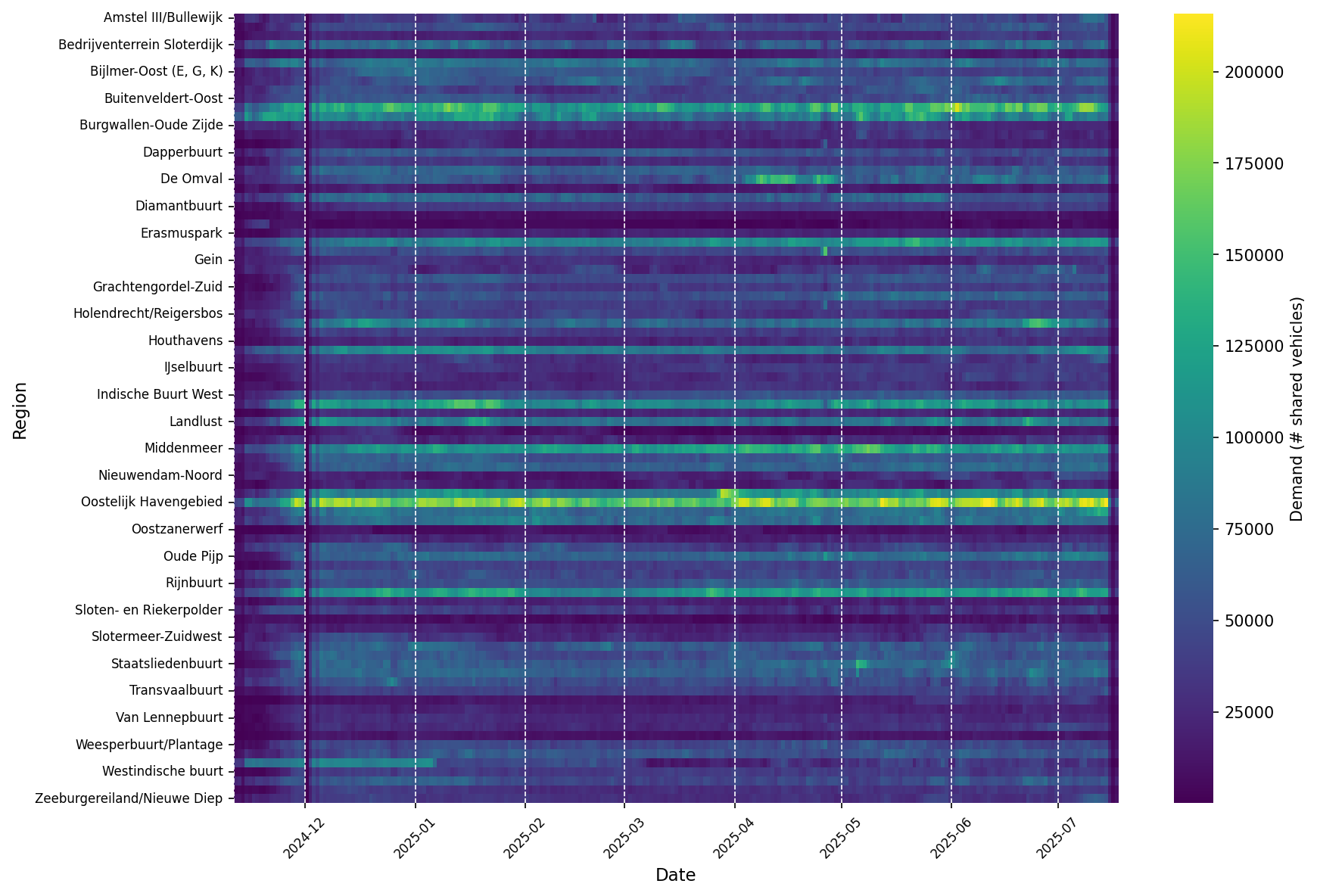}
    \caption{Daily demand across different districts in the Rotterdam dataset. Each row represents a district, and each column corresponds to a day in the observation period. Color intensity reflects the total daily demand (aggregated from minute-level data), with warmer colors indicating higher activity.}
    \label{fig:daily_heatmap}
\end{figure}

\begin{figure}[h!t]
    \centering
    \includegraphics[width=.9\columnwidth]{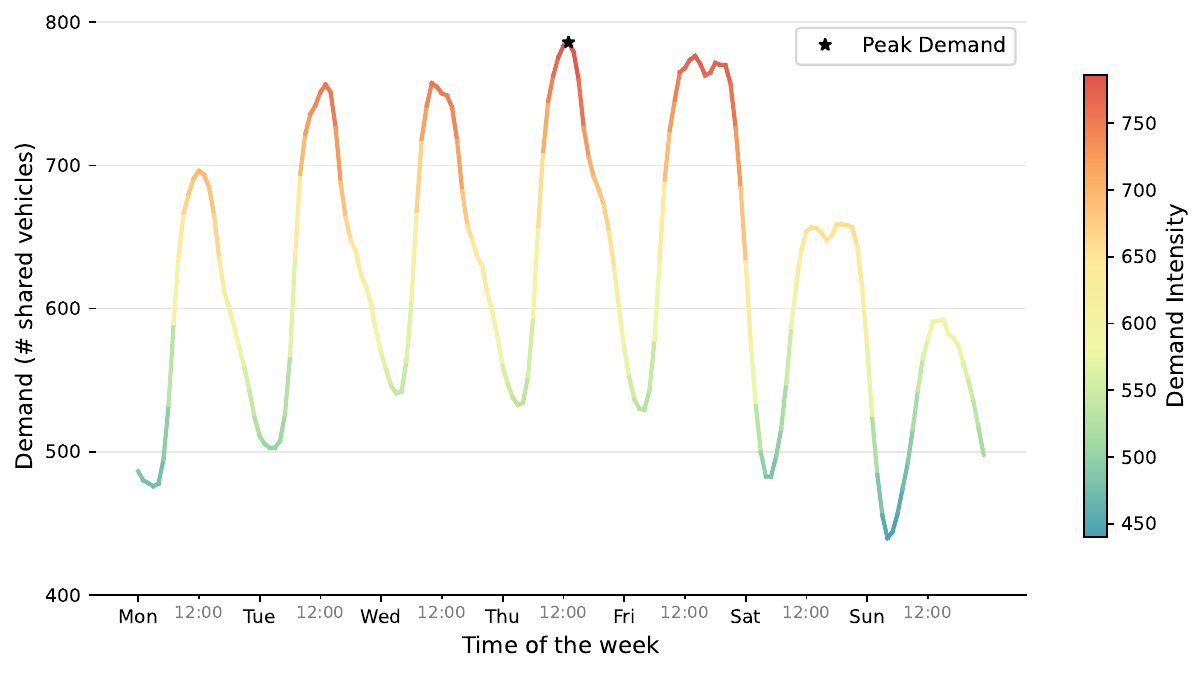}
    \caption{Average hourly demand across the week in Rotterdam Centrum. The coloring indicates the relative demand intensity and, along with the peaks, aids in highlighting daily and weekly usage patterns.}
    \label{fig:HoW}
\end{figure}

Moreover each one of these timeseries may exhibit intricate mobility patterns, which will deviate from the "expected" mobility patterns. Figure \ref{fig:daily_heatmap} exemplifies this statement by showing the total daily mobility demand across the different Rotterdam districts. t is evident that, while some districts have mostly high demand, often exceeding 200K shared vehicles (e.g., Oostelijk Havengebied) while others have sparse to almost non-existent demand (e.g., Botlek). In addition, Figure \ref{fig:HoW} visualizes the weekly demand profile of a major district in Rotterdam (namely the Rotterdam Centrum) by showing the interaction between hour-of-day and day-of-week. The figure reveals consistent commuting patterns, with the highest and most prolonged peak occurring on Fridays, likely due to the convergence of work and leisure activities.

\begin{figure}[h!t]
    \centering
    \includegraphics[width=\columnwidth]{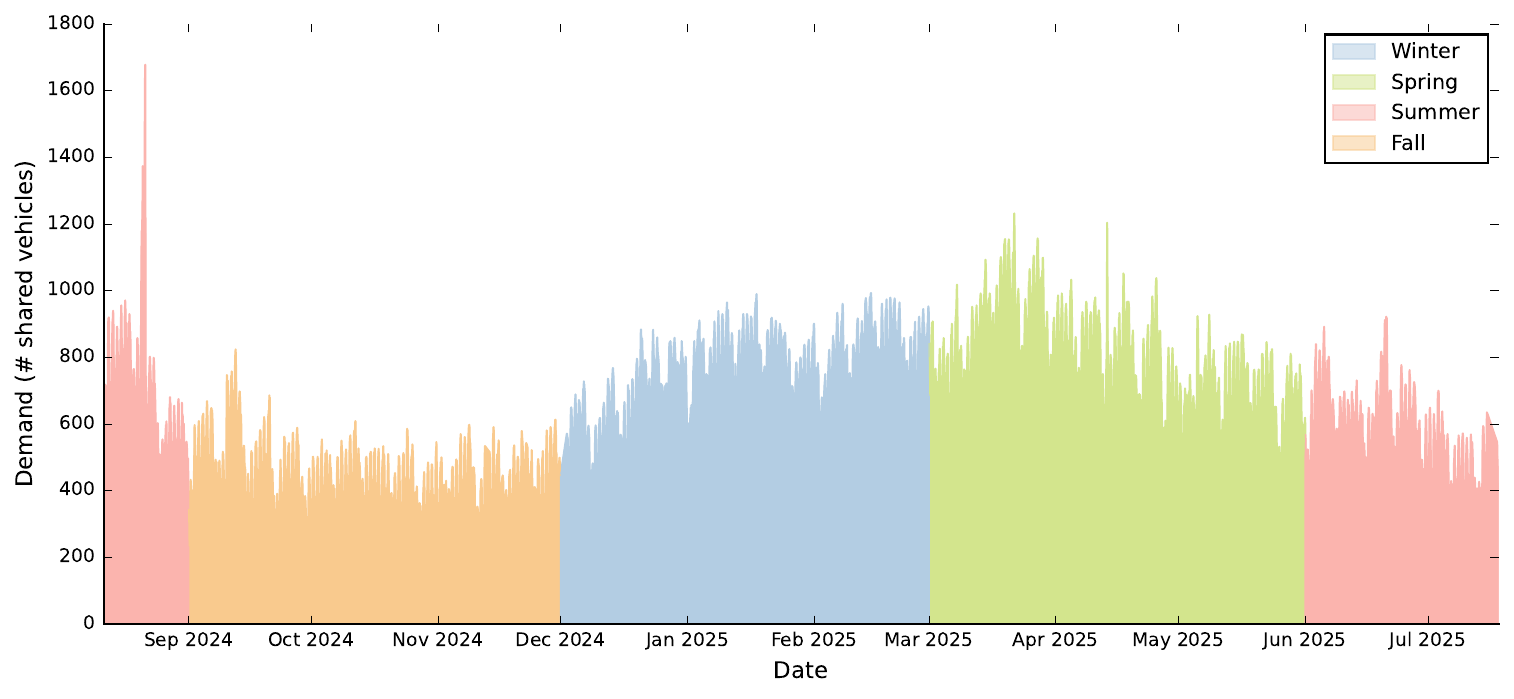}
    \caption{Temporal evolution of the shared mobility demand (in hourly basis) in Rotterdam Centrum. Seasons are color-coded to highlight the different behaviors. During fall and winter seasons, demand patterns are more stable, whereas during spring and summer seasons, short-term spikes are observed often at “random” periods.}
    \label{fig:evolve}
\end{figure}

These patterns are further highlighted in Figure \ref{fig:evolve}, which visualizes the seasonal demand evolution in the Rotterdam Centrum district. One can easily distinguish the commute patterns within a workday (concentrated peaks around midday) and the leisure use on the weekends with a more spread-out peak demand around the middle of the days.

\subsection{The MoDE-Boost Framework Pipeline}\label{sec:methodology}
In this section, we present our MoDE-Boost framework, designed to efficiently address the SMDF problem defined in Section \ref{sec:problemdef}. The framework is built by accommodating both classification and regression objectives, depending on the nature of the downstream task, as follows:
\begin{itemize}
    \item Classification-based MoDE-Boost (C-MoDE-Boost): The model is trained to classify future demand into discrete ordinal categories (e.g., Low, Medium, or High) based on demand thresholds computed from historical distribution statistics.
    \item Regression-based MoDE-Boost (R-MoDE-Boost): The model is trained to predict the exact numerical value of future demand.
\end{itemize}

Particularly, given an urban area partitioned into a number of regions (recall the r-SMDF problem) or including a number of points/stations (the p-SMDF problem, respectively), and a multi-step forecasting objective at multiple future horizons (e.g., 5 min., 15 min., 30 min., and 60 min.), our framework centers around a model trained jointly across all regions (or points, respectively) of the area. This unified approach is designed to leverage shared spatio-temporal regularities across regions / points while retaining district-specific nuances via the inclusion of the region / point identifier as an input feature. This identifier is encoded and integrated into the feature space, allowing the model to implicitly learn distinct behavioral patterns per region / point.

\begin{figure}[t]
    \centering
    \includegraphics[width=\linewidth]{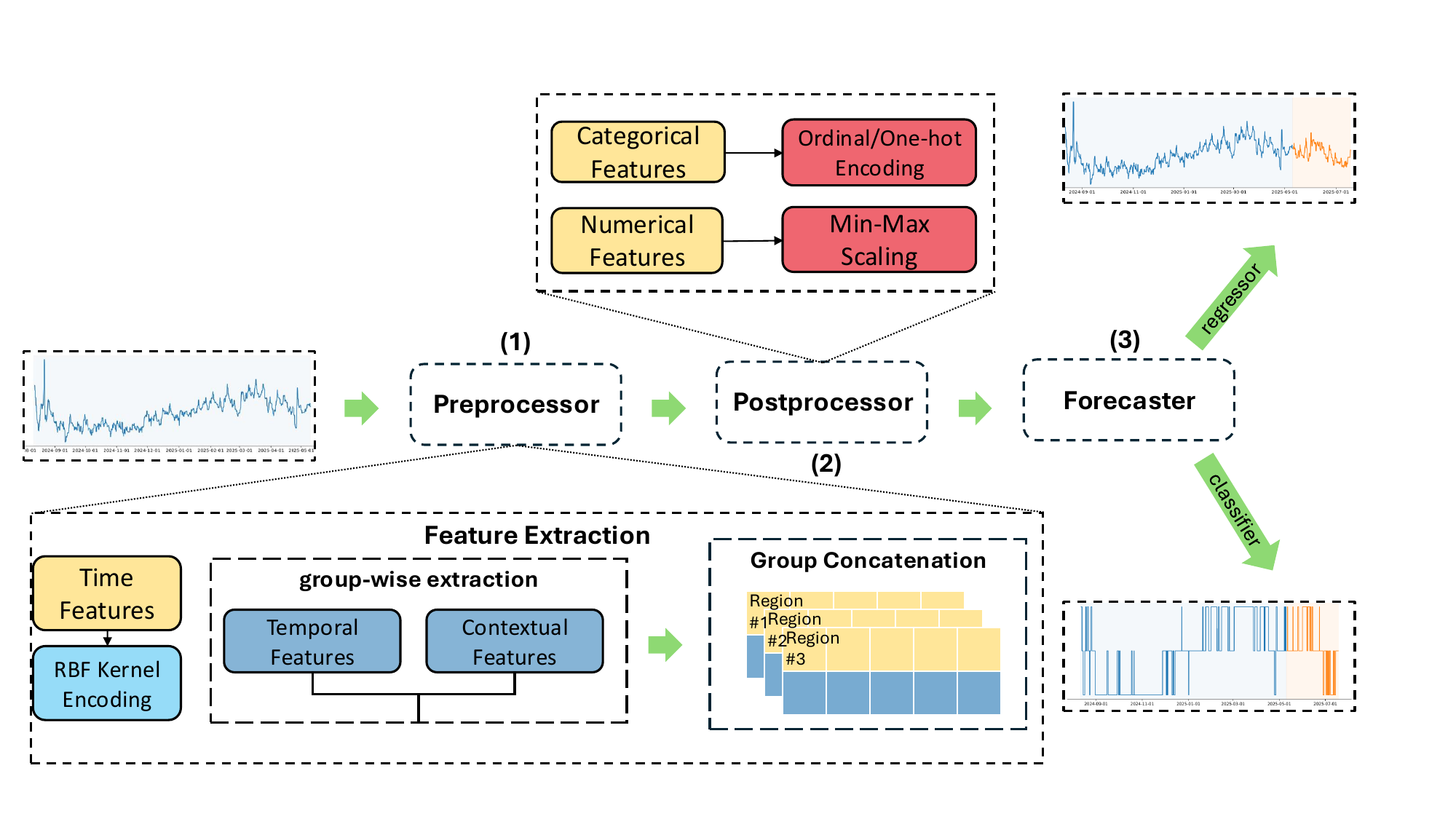}
    \caption{Overview of our proposed SMDF framework: 
        (1) a preprocessing stage creates enriched per‑region timeseries; 
        (2) postprocessing normalizes and encodes features while preserving inter‑region demand ratios; 
        (3) an XGBoost predictor produces the demand forecasts.}
    \label{fig:problem-viz}
\end{figure}

As illustrated in Figure \ref{fig:problem-viz}, our methodology consists of three steps, as follows:
\begin{enumerate}
    \item \textbf{Preprocessor.}  Performs the bulk of feature engineering: the raw data are partitioned into entity-specific timeseries, temporal and contextual attributes are extracted for each spatial entity, and the enriched feature groups are merged to obtain an augmented dataset.
    \item \textbf{Postprocessor.}  Normalizes numerical variables and encodes categorical ones.  Normalization is applied jointly across all regions so that relative demand magnitudes (e.g., $ D_A > D_B$) are preserved.
    \item \textbf{Forecaster.}  Supplies the normalized feature matrix to an XGBoost model (regressor or classifier), which generates the final demand forecasts.
\end{enumerate}
    
\subsection{Feature Extraction}\label{sec:feature-extraction}

To accurately capture the spatio-temporal dynamics underlying shared micro-mobility demand, extensive feature engineering is performed as a preprocessing step. Although the proposed model architecture is designed to process inputs from multiple spatial entities simultaneously as already discussed in Section \ref{sec:methodology}, feature extraction is conducted independently for each one to preserve the locality and heterogeneity of mobility patterns. This approach ensures that the unique demand characteristics and flow dynamics intrinsic to each area are not diluted by global averaging effects. Apart from the base features about the demand value and the corresponding spatial entity, the engineered features are structured to capture key dimensions: 
(i) temporal dynamics, including recurring patterns such as hourly, daily and weekly seasonality; and
(ii) contextual features, incorporating exogenous variables such as public holidays, or other social factors (e.g., school/work-hours) where applicable. This composite representation enhances the model's ability to learn fine-grained, localized demand signals while preserving global consistency across the shared mobility network.

The features used in our methodology are summarized in Table \ref{tab:feature_description} and presented in detail in the paragraphs that follow.

\begin{table*}[t]
    \fontsize{16}{12}\selectfont
    \renewcommand{\arraystretch}{1.8}
    \centering
    \caption{Summary table of the extracted features used in our approach}
    \label{tab:feature_description}
    \resizebox{\textwidth}{!}{%
    \begin{tabular}{@{}llll@{}}
        \toprule
        \textbf{Category} & \textbf{Type} & \textbf{Feature Name} & \textbf{Description} \\
        \midrule
        \multirow{2}{*}{Base} 
            & ID & \texttt{region (station)} & Spatial entity identifier \\
            & value & \texttt{demand} & Raw demand value \\ \midrule
        \multirow{8}{*}{Temporal} 
            & Lagged & \texttt{demand\_lag\_*} & Lagged demand values $n$ minutes ago; $n\in\{1, 5, 60, 1440\}$\\ \cmidrule(l){3-4}
            & \multirow{2}{*}{Rolling} 
            & \texttt{demand\_rolling\_mean\_*} & Rolling mean of demand over past $n$ time steps; $n\in\{5, 10, 60, 1440\}$ \\
            & & \texttt{demand\_rolling\_max\_*} & Rolling maximum of demand over past $n$ time steps; $n\in\{5, 10, 60, 1440\}$ \\ \cmidrule(l){3-4}
            & EWMA & \texttt{demand\_ewm\_*} & \makecell[l]{Exponentially weighted moving average of the demand value over \\past $n$ time steps; $n\in\{5, 10, 60, 1440\}$} \\ \cmidrule(l){3-4}
            & \multirow{3}{*}{Statistical} 
            & \texttt{rolling\_demand\_cv} & Coefficient of variation (std/mean) over the last hour/day \\
            & & \texttt{demand\_magnitude} & Category of the demand volume regime, derived from quantile-based bins \\
            & & \texttt{demand\_adjusted} & \makecell[l]{Ordinally scaled demand that normalizes raw demand by applying\\ a level‑specific multiplier} \\ \cmidrule(l){3-4}
            & Fourier & \texttt{fourier\_demand} & Fourier-transformed component to capture periodic patterns in demand data \\ 
        \midrule
        \multirow{5}{*}{Contextual} 
            & \multirow{4}{*}{Calendar} & \texttt{hour} & Cyclical encoding of hour of day [0..23] \\
            &  & \texttt{day} & Cyclical  encoding of day of week [0..6] \\
            &  & \texttt{month} & Cyclical encoding of month of year [1..12] \\ 
            &  & \texttt{quarter} & Cyclical encoding of quarter of year [1..4] \\ 
            \cmidrule(l){2-4}
            & Holiday & \texttt{is\_holiday\_period} & Whether a date is a holiday or the day before/after a holiday [yes, no] \\
        \bottomrule
    \end{tabular}%
    }
\end{table*}

\subsubsection{Temporal Features}

\paragraph{Lagged features}
Lagged features in the dataset are designed to capture demand dynamics across multiple time scales. This category encodes temporal dependencies by shifting historical demand values backward in time for each district. Formally, for a demand value $D^A_t$ observed at time $t$ in spatial entity $A$, a lagged feature with offset $\tau\in\mathbb{N}$ is defined as $D^A_{t-\tau}$. To capture both immediate and slightly delayed effects, we extract lagged values at offsets of 1 min., 5 min., 15 min., and 60 min., as well as the corresponding observations from the previous day (i.e., 1440 min.). This setup allows the model to incorporate recent system memory, enabling it to detect recurrent patterns and local autocorrelations that influence future demand.

\paragraph{Rolling features}
Rolling (windowed) statistics capture short‑term dynamics by aggregating the historical demand vector $\mathbf{X}^{A}_{1:t}$ over fixed‑length sliding windows.  
For a window length $w\in\{5,\,10,\,60,\,1440\}$ minutes (the latter corresponding to a full day), we compute

\[
\mu^{A}_{t,w}= \operatorname{mean}\!\bigl(\mathbf{X}^{A}_{t-w+1:t}\bigr),\qquad
\omega^{A}_{t,w}= \operatorname{max}\!\bigl(\mathbf{X}^{A}_{t-w+1:t}\bigr),
\]

where $\mu^{A}_{t,w}$ denotes the \textit{rolling mean} (local central tendency) and $\omega^{A}_{t,w}$ the \textit{rolling maximum} (peak demand) for spatial entity $A$ at time $t$.  
These windowed aggregates smooth noisy fluctuations while preserving the underlying temporal structure.  
Short windows (e.g., $w=5$ min.) highlight rapid changes such as event‑driven surges, whereas long windows (e.g., $w=1440$ min.) capture broader patterns like the buildup toward daily peak hours.  
In the present framework, only the above aggregate functions are implemented; additional statistics (e.g., rolling standard deviation, minimum, kurtosis) can be incorporated straightforwardly to quantify local volatility when needed.

\paragraph{Exponentially Weighted features}
Exponentially weighted statistics give larger influence to the most recent observations while still retaining information from the more distant past.  
The Exponentially Weighted Moving Average (EWMA) for spatial entity $A$ at time $t$ is defined recursively as  

\[
\operatorname{EWMA}^{A}_{t}
    = \alpha\, D^{A}_{t} + (1-\alpha)\,\operatorname{EWMA}^{A}_{t-1},
    \qquad \alpha\in(0,1],
\]

where $\alpha$ is the decay factor that controls how quickly past observations are discounted.  
Unlike a fixed‑window mean, EWMA adapts automatically to sudden demand shifts (e.g., weather‑induced spikes or large public events) while preserving a long‑term context.  
In practice we relate $\alpha$ to an effective window size $N$ via

\[
\alpha=\frac{2}{N+1},
\qquad N\in\{5,\,10,\,60,\,1440\},
\]

so that $N$ ranges from 5 minutes to a full day (1440 minutes).

\paragraph{Statistical features}
Demand magnitudes can differ dramatically across districts, which may impair the learning of a single spatio‑temporal model.  
To mitigate this disparity we introduce an \textit{Adjusted Demand} transformation.  
First, the raw demand $D^{A}_{t}$ is assigned to one of $L$ ordinal levels (e.g., Low, Medium, High) based on quantile thresholds.  
Each level $\ell$ receives a scaling factor $s_{\ell}$; in our implementation the central level (Medium) has $s_{\ell}=1$, the adjacent levels are multiplied by $s$ or $1/s$ (here $s=2$), and more extreme levels use higher powers of $s$.  
The adjusted demand is then calculated as:
\[
\widetilde{D}^{A}_{t} = \gamma\!\bigl(D^{A}_{t}\bigr)\; D^{A}_{t},
\]
where $\gamma\!\bigl(D^{A}_{t}\bigr)$ denotes the multiplier for the corresponding quantile level.  
This operation preserves the semantic ordering of demand while compressing inter‑district variance.  
Complementarily, we quantify local volatility through the \textit{coefficient of variation} computed on a rolling window of length $w$:

\[
\operatorname{CV}^{A}_{t,w} = \frac{\sigma^{A}_{t,w}}{\mu^{A}_{t,w}},
\qquad
\sigma^{A}_{t,w}= \operatorname{std}\!\bigl(\mathbf{X}^{A}_{t-w+1:t}\bigr),
\;\;
\mu^{A}_{t,w}= \operatorname{mean}\!\bigl(\mathbf{X}^{A}_{t-w+1:t}\bigr).
\]

This combination of ordinal scaling and variability assessment yields a more homogeneous feature space for downstream spatio‑temporal modeling.

\paragraph{Fourier‑based (spectral) features}
Urban micro-mobility demand exhibits strong periodicities (daily commuting cycles, weekly event rhythms).  
Fourier decomposition isolates these cyclic components by projecting the raw signal onto orthogonal sine and cosine bases.  
For a chosen set of $K$ low‑frequency harmonics we compute  

\[
\begin{aligned}
a^{A}_{k} &= \frac{2}{P}\sum_{t=1}^{P} D^{A}_{t}\,\cos\!\Bigl(2\pi\frac{t}{P}k\Bigr),\\
b^{A}_{k} &= \frac{2}{P}\sum_{t=1}^{P} D^{A}_{t}\,\sin\!\Bigl(2\pi\frac{t}{P}k\Bigr),
\end{aligned}
\qquad k=1,\dots,K,
\]
where $P$ is the period of interest (e.g., $P=1440$ for a daily cycle).  
The reconstructed (denoised) series using only the dominant harmonics is  

\[
\widehat{D}^{A}_{t}
    = \sum_{k=1}^{K}\Bigl(a^{A}_{k}\cos\!\bigl(2\pi \tfrac{k}{P}t\bigr)
    + b^{A}_{k}\sin\!\bigl(2\pi \tfrac{k}{P}t\bigr)\Bigr).
\]

These harmonic coefficients $a^{A}_{k}$ and $b^{A}_{k}$ constitute compact frequency‑domain descriptors that complement traditional time‑domain predictors (lagged values, rolling statistics).  
By summarizing the long‑range regularities of demand while filtering short‑term noise, Fourier‑based features reveal latent periodic signals that are otherwise obscured, thereby improving model generalization and the detection of stable temporal trends.

\subsubsection{Contextual Features}
\textit{Calendar} features are designed to capture recurring temporal patterns in mobility demand, reflecting human behavioral rhythms and urban activity cycles across multiple time scales. Starting with the most granular, minute-of-hour (0–59) adds fine-grained temporal resolution, enabling detection of intra-hour dynamics, particularly useful in high-frequency data scenarios. Hour-of-day (0–23) encodes diurnal fluctuations, such as commute peaks. Day-of-week (0–6, Monday as 0) captures weekly variations, distinguishing between structured weekday routines and more variable weekend behavior. Month-of-year (0–11, January as 0) models seasonal trends, accounting for climate-driven or tourism-related fluctuations in demand. Quarter-of-year (1-4, Winter as 1) captures seasonal shifts in mobility patterns (demand is expected during winter, more during summer). Finally, we also consider for holiday events and days around it (1 day before and after) using the \textit{is\_holiday\_period} feature.

\subsection{Demand Classification}\label{sec:classif}
Especially for the C-MoDE-Boost version of our framework that addresses the demand value forecasting as a classification task, the definition of the classes is based on the relative deviation from the maximum observed daily demand across the dataset. Let the “peak” daily demand be scaled to 100 and let $d\in(0,50)$ represent a symmetric tolerance threshold. We define medium demand the values within the interval $[50-d,50+d)$ and low (high) demand the values below (above) $50-d$ ($50 + d$, respectively). In our experiments, we set $d = 20$, resulting in three equally sized quantile-based intervals. This threshold provides a balanced distribution of samples across demand levels, which facilitates robust classification.
\section{Experimental Study}\label{sec:experimentalstudy}

This section describes the data and preprocessing steps used in our experimental study (Section \ref{sec:setup}) as well as our findings regarding the region-based and the point-based variants of the SMDF problem (Section \ref{sec:rSMDF} and Section \ref{sec:pSMDF}, respectively).
The experiments were conducted on a server with an AMD EPYC 64-Core CPU and an NVIDIA A100 GPU with 40GB of memory. In addition and in order to assess our argument that our framework is Edge-ready, in Section 4.3 we present a set of experiments performed at the Edge. In particular, we used a Raspberry Pi 5 (8-core ARM Cortex-A76, 16 GB RAM), a widely accessible and energy-efficient edge device commonly used in IoT and smart city applications.

\subsection{Experimental Setup}\label{sec:setup}

\subsubsection{Dataset Description}\label{sec:data-desc}

To assess the performance of our proposed MoDE-Boost framework we employ two different kinds of datasets that correspond to the two modeling settings introduced earlier (region- vs. point-based problem).

For the \textbf{r}-SMDF problem, we leverage real‑time shared micro‑mobility data provided by \textit{Deelfiets Nederland} via an open API\footnote{\url{https://api.deelfietsdashboard.nl/dashboard-api/public/vehicles_in_public_space}}. 
Every 60 seconds, the API returns the GPS position (WGS‑84), vehicle type (bicycle, e‑bike, e‑scooter) and operator name of all active units. 
For the purposes of our experimental study, we extracted mobility traces from three major metropolitan areas (Amsterdam, Rotterdam, and The Hague). 
Timestamps were floored to the minute and appended to the series and the number of active vehicles was aggregated per $<$spatial entity, minute$>$ pair. 
The resulting $<$id, timestamp, value$>$ tuples feed the region-based variation of the MoDE-Boost framework, presented in Figure \ref{fig:problem-viz}.

On the other hand, for \textbf{p}-SMDF problem, we experiment with two publicly available, de‑facto benchmark datasets, namely the Citi Bike (NYC) – the official NYC system‑wide trip archive\footnote{\url{https://citibikenyc.com/system-data}}, and Divvy (Chicago) – the corresponding Chicago bike‑share dataset\footnote{\url{https://divvybikes.com/system-data}}. 
For both datasets, we restrict the analysis to the period between January 1st, 2021 and June 1st, 2022, so that our experiments are directly comparable with the related work \cite{FengLiuASTN}.
A uniform cleaning pipeline is then applied: records with missing fields are discarded, trips longer than 24h, implausibly short round‑trips and stations that average fewer than three rentals per day are excluded. After preprocessing, the CitiNYC comprises \(\lvert\mathcal{V}\rvert = 1{,}374\) stations and a total of \(35{,}160{,}150\) valid trips. The Divvy dataset is processed identically, yielding a comparable set of  \(\lvert\mathcal{V}\rvert = 1{,}007\) stations and a total of \(34{,}669{,}620\) valid trips.
A description of the datasets used in our paper along their main features is presented in Table \ref{tab:data-desc}.

\begin{table*}[t]
    \fontsize{12}{10}\selectfont
    \renewcommand{\arraystretch}{1.8}
    \centering
    \caption{Summary table of the datasets used in our experimental study.}
    \label{tab:data-desc}
    \resizebox{\textwidth}{!}{%
    \begin{tabular}{@{}llccc@{}}
        \toprule
        Variation                       & Dataset         & Time Span               & \# Records & \makecell[c]{\# Spatial Entities\\(regions or points)} \\ \midrule
        \multirow{3}{*}{Region-based}   & Rotterdam       & 2024-08-11 - 2025-07-18 & 7,561,051  & 13                  \\
                                        & Amsterdam       & 2024-11-11 - 2025-07-18 & 32,925,151 & 88                  \\
                                        & The Hague       & 2024-12-11 - 2025-07-18 & 13,099,811 & 46                  \\ \cmidrule(l){1-5} 
        \multirow{2}{*}{Point-based}    & CitiNYC         & 2021-01-01 - 2022-06-01 & 36,323,801 & 1374                \\
                                        & Divvy           & 2021-01-01 - 2022-06-01 & 7,104,591  & 1007                \\ 
        \bottomrule
        \end{tabular}%
    }
\end{table*}

To validate our model, we employ a chronological split of the datasets using a 70:20:10 ratio for the train, validation, and test set, respectively. 
This method preserves the temporal integrity of the data, preventing leakage of future information and ensuring a realistic evaluation of model performance. 
In preparation for training C-MoDE-Boost, the classifier variant of our architecture, an additional preprocessing step was applied to transform the continuous demand values into discrete categories, as it was described in Section \ref{sec:classif}

\subsubsection{Hyperparameter Optimization}\label{sec:hyperopt}
To improve the predictive performance of our framework, we employ a Bayesian‑optimization‑based hyper‑parameter search \cite{optuna2019}, a strategy that is well‑suited to the costly training cycles inherent to our forecasting models.
By iteratively constructing a probabilistic surrogate of the validation objective, the optimizer efficiently balances exploration of the hyper‑parameter space with exploitation of promising regions, thereby converging to high‑quality configurations with far fewer evaluations.
The search process is guided by a Tree‑structured Parzen Estimator (TPE) sampler, which discriminates between favorable and unfavorable trial distributions \cite{Bergstra2011TPESampler}, and its median‑based pruning mechanism, which discards under‑performing trials early to conserve computational resources.
A two‑phase, coarse‑to‑fine refinement is adopted: an initial broad search identified a promising sub‑space, after which a subsequent narrower search centered on the best trial fine‑tuned the parameters.
This hierarchical approach yields robust, computationally efficient hyper‑parameter settings that support scalable, city‑wide demand forecasting, and render the tuned MoDE-Boost model suitable for deployment on edge devices.

\subsection{Experimental Results}\label{sec:exp-results}
In this section, we assess the performance of our model when addressing the two variants of the problem at hand (i.e., r-SMDF in Section \ref{sec:rSMDF} and p-SMDF in Section \ref{sec:pSMDF}). 

\subsubsection{Assessing the region-based variant of the MoDE-Boost framework}\label{sec:rSMDF}
The first set of experiments aims to benchmark MoDE-Boost using the "Region" group of  datasets described in Table \ref{tab:data-desc} to (i) quantify training and inference latency and (ii) evaluate its predictive accuracy across all four forecasting horizons. The next set of experiments is focused on comparing our approach with four benchmark models, namely, a Historical Average model, a Seasonal Naïve with daily seasonality, a Simple Exponential Smoothing model \cite{HoltWintersSES}, and Croston's method \cite{Croston1972}, an advanced model designed for intermittent demand.
The third set of experiments aims to compare our region-based approach with: (i) a specialized variation, where instead of training a single model across all spatial entities, we train and fine-tune a single model for each entity and (ii) a Generative AI-based approach using the TimeGPT FM.
Collectively, these results demonstrate that the MoDE-Boost framework (i) consistently surpasses established baselines in predictive accuracy, (ii) scales efficiently in both training and deployment

Table \ref{tab:timing-summary} summarizes the training and inference times required for our models across four forecasting horizons (5, 15, 30, and 60 min.). Training time refers to the total duration required to train a single dataset-horizon model, while inference time denotes the average time taken to produce a single prediction at a specific horizon. The results indicate that both variations of our model (R-MoDE-Boost for regression and C-MoDE-Boost for classification)  are highly efficient: training each model typically takes 15.5 sec on average or less in many cases, while inference latency is remarkably low – on average 0.37 $\mathrm{\mu}$sec. This highlights the models’ potential for real-time deployment in intelligent transportation systems and other urban applications with stringent latency constraints.

\begin{table*}[h!]
    \fontsize{14}{8}\selectfont
    \centering
    \caption{Training and inference times of the MoDE-Boost model per dataset, model, and forecasting horizon. The sizes of the training and test sets of the datasets are provided for reference of the training and inference times, respectively.}
    \label{tab:timing-summary}
    \renewcommand{\arraystretch}{1.5}
    {\setlength{\tabcolsep}{3pt}
     \scriptsize
     \resizebox{\linewidth}{!}{%
        \begin{tabular}{l l c c c c c}
        \toprule
        \textbf{Dataset} &
        \textbf{Model} &
        \textbf{\makecell[t]{Horizon\\(min)}} &
        \textbf{\makecell[t]{Train Size\\(records)}} &
        \textbf{\makecell[t]{Training Time (sec)\\(mean ± std)}} &
        \textbf{\makecell[t]{Test Size\\(records)}} &
        \textbf{\makecell[t]{Inference\\Time (µsec)}} \\
        \midrule
        \multirow{8}{*}{Amsterdam} &
        \multirow{4}{*}{Regressor} &
        5  & 24.8M & 29.83 ± 7.54 & 2.7M & 0.34 \\
        & & 15 & 24.8M & 8.59 ± 1.19  & 2.7M & 0.32 \\
        & & 30 & 24.7M & 15.15 ± 2.73 & 2.7M & 0.32 \\
        & & 60 & 24.7M & 13.63 ± 1.97 & 2.7M & 0.33 \\ \cmidrule(lr){3-7}
        & \multirow{4}{*}{Classifier} &
        5  & 24.8M & 34.43 ± 0.55 & 2.7M & 0.35 \\
        & & 15 & 24.8M & 48.86 ± 0.34 & 2.7M & 0.35 \\
        & & 30 & 24.7M & 15.10 ± 0.46 & 2.7M & 0.29 \\
        & & 60 & 24.7M & 45.28 ± 0.15 & 2.7M & 0.33 \\[2pt]
        \midrule
        \multirow{8}{*}{The Hague} &
        \multirow{4}{*}{Regressor} &
        5  & 9.3M & 4.95 ± 0.10 & 1M & 0.36 \\
        & & 15 & 9.3M & 9.68 ± 2.30 & 1M & 0.35 \\
        & & 30 & 9.3M & 5.38 ± 1.10 & 1M & 0.33 \\
        & & 60 & 9.3M & 5.09 ± 0.50 & 1M & 0.34 \\ \cmidrule(lr){3-7}
        & \multirow{4}{*}{Classifier} &
        5  & 9.3M & 4.79 ± 0.07 & 1M & 0.29 \\
        & & 15 & 9.3M & 16.25 ± 0.68 & 1M & 0.35 \\
        & & 30 & 9.3M & 12.99 ± 0.30 & 1M & 0.32 \\
        & & 60 & 9.3M & 24.12 ± 0.49 & 1M & 0.36 \\[2pt]
        \midrule
        \multirow{8}{*}{Rotterdam} &
        \multirow{4}{*}{Regressor} &
        5  & 5M & 2.97 ± 0.71 & 558K & 0.34 \\
        & & 15 & 5M & 4.85 ± 0.74 & 558K & 0.43 \\
        & & 30 & 5M & 3.29 ± 0.21 & 558K & 0.47 \\
        & & 60 & 5M & 8.65 ± 4.46 & 558K & 0.42 \\ \cmidrule(lr){3-7}
        & \multirow{4}{*}{Classifier} &
        5  & 5M & 12.05 ± 0.16 & 558K & 0.50 \\
        & & 15 & 5M & 10.42 ± 0.72 & 558K & 0.48 \\
        & & 30 & 5M & 26.37 ± 0.16 & 558K & 0.52 \\
        & & 60 & 5M & 9.90 ± 0.41  & 558K & 0.47 \\[2pt]
        \bottomrule
    \end{tabular}
     }
    }
\end{table*}

Regarding the prediction accuracy of our model, Tables \ref{tab:performance-reg} and \ref{tab:performance-clf} summarize the predictive performance of the proposed spatio‑temporal forecasting models across four forecasting horizons (5 min., 15 min., 30 min., and 60 min.). For the regression task, we report the well-known RMSE and MAE quality metrics, whereas for the classification task, we present F1‑score and Accuracy. These complementary metrics jointly capture the models’ predictive performance, robustness to outliers, and ability to discriminate between demand levels across spatial entities that differ in population density and temporal dynamics. Overall, our model is shown to achieve consistently high classification scores in the three Dutch cities across different horizons.

\begin{table}[h!]
    \centering
    \renewcommand{\arraystretch}{1.2}
    \caption{Performance metrics of r-MoDE-Boost across four forecasting horizons.}
    \label{tab:performance-reg}
    \fontsize{8}{12}\selectfont
    \begin{tabular}{@{}lcccccccc@{}}
        \toprule
        \multirow{2}{*}{\textbf{Dataset}} &
        \multicolumn{2}{c}{\textbf{5 min}} &
        \multicolumn{2}{c}{\textbf{15 min}} &
        \multicolumn{2}{c}{\textbf{30 min}} &
        \multicolumn{2}{c}{\textbf{60 min}} \\
        \cmidrule(l){2-3} \cmidrule(l){4-5} \cmidrule(l){6-7} \cmidrule(l){8-9}
        & \textbf{MAE} & \textbf{RMSE} 
        & \textbf{MAE} & \textbf{RMSE} 
        & \textbf{MAE} & \textbf{RMSE} 
        & \textbf{MAE} & \textbf{RMSE} \\
        \midrule
        Amsterdam & 0.95 & 1.37 & 1.28 & 1.78 & 1.59 & 2.12 & 2.02 & 2.66 \\
        The Hague & 1.05 & 1.52 & 1.25 & 1.78 & 1.44 & 2.09 & 1.85 & 2.61 \\
        Rotterdam & 4.70 & 6.33 & 5.74 & 7.80 & 7.14 & 9.80 & 9.15 & 12.05 \\
        \bottomrule
    \end{tabular}%
\end{table}

\begin{table}[h!]
    \fontsize{8}{12}\selectfont
    \caption{Performance metrics of c-MoDE-Boost across four forecasting horizons.}
    \label{tab:performance-clf}
    \centering
    \renewcommand{\arraystretch}{1.2}
    \begin{tabular}{@{}ccccccccc@{}}
        \toprule
        \multirow{2}{*}{\textbf{Dataset}} &
          \multicolumn{2}{c}{\textbf{5 min}} &
          \multicolumn{2}{c}{\textbf{15 min}} &
          \multicolumn{2}{c}{\textbf{30 min}} &
          \multicolumn{2}{c}{\textbf{60 min}} \\ \cmidrule(l){2-3} \cmidrule(l){4-5} \cmidrule(l){6-7} \cmidrule(l){8-9}   
         &
          \textbf{F1} & \textbf{Accuracy} &
          \textbf{F1} & \textbf{Accuracy} &
          \textbf{F1} & \textbf{Accuracy} &
          \textbf{F1} & \textbf{Accuracy} \\ \midrule
        Amsterdam & 0.88 & 0.86 & 0.85 & 0.84 & 0.83 & 0.81 & 0.79 & 0.76 \\
        The Hague & 0.90 & 0.84 & 0.88 & 0.81 & 0.87 & 0.79 & 0.85 & 0.76 \\
        Rotterdam & 0.95 & 0.90 & 0.93 & 0.84 & 0.92 & 0.82 & 0.89 & 0.77 \\ 
        \bottomrule
    \end{tabular}
\end{table}

We also demonstrate the predictive capabilities of our approach by comparing MoDE-Boost with the benchmark models as well as the TimeGPT Foundation Model \cite{garza2023timegpt1}. Table \ref{tab:perf-benchmark} provides a comprehensive performance comparison of r-MoDE-Boost against related work on the Rotterdam shared mobility dataset. Our approach outperforms the three baselines models by a relatively great margin and slightly outperforming the other two models. Our model attains the lowest RMSE in nine districts  and the second‑lowest RMSE in four additional districts. In terms of MAE, r‑MoDE-Boost is the best performer in eight districts. These complementary error patterns suggest that our model excels at modeling routine demand fluctuations while still capturing enough variance of the timeseries.

The paired predictions of r‑MoDE-Boost and the TimeGPT baseline were subjected to a paired paired Student’s \textit{t}-test ($p=2.47\times10^{-4}$) and a Wilcoxon Signed‑Rank test ($p=4\times10^{-8}$), both well below the $\alpha=0.01$ significance threshold. Consequently, we reject the null hypothesis of equal performance and conclude that the observed improvement is not attributable to random variation in the test split. r-MoDE-Boost reduces the mean absolute error from 12.31 to 8.35 (a gain of 3.96 units, $\approx32\%$).

\begin{table*}[h!]
    \fontsize{14}{6}\selectfont
    \centering
    \caption{Average city-level performance comparison for the 60 min. horizon (the lowest error in bold type).}
    {\setlength{\tabcolsep}{4pt}
     \scriptsize
     \resizebox{\textwidth}{!}{%
     \begin{tabular}{lccccccccccc}
     \toprule
        \multicolumn{1}{l}{\multirow{2}{*}{\textbf{Dataset}}} &
        \multicolumn{2}{c}{\textbf{HA}} &
        \multicolumn{2}{c}{\textbf{SN}} &
        \multicolumn{2}{c}{\textbf{SES}} &
        \multicolumn{2}{c}{\textbf{Croston}} &
        \multicolumn{2}{c}{\textbf{r-MoDE-Boost}} \\
        \cmidrule(l){2-3}\cmidrule(l){4-5}\cmidrule(l){6-7}
        \cmidrule(l){8-9}\cmidrule(l){10-11}
        & MAE & RMSE & MAE & RMSE & MAE & RMSE & MAE & RMSE & MAE & RMSE \\ \midrule
        Amsterdam &
        \makecell{5.88\\($\pm$ 3.04)} & \makecell{7.21\\($\pm$ 3.66)} &
        \makecell{5.35\\($\pm$ 2.35)} & \makecell{6.86\\($\pm$ 3.07)} &
        \makecell{2.29\\($\pm$ 0.98)} & \makecell{3.08\\($\pm$ 1.28)} &
        \makecell{2.29\\($\pm$ 0.98)} & \makecell{3.08\\($\pm$ 1.28)} &
        \makecell{\textbf{2.06}\\($\pm$ 0.86)} & \makecell{\textbf{2.78}\\($\pm$ 1.12)} \\ \cmidrule{2-11}
        The Hague &
        \makecell{9.06\\($\pm$ 7.97)} & \makecell{10.45\\($\pm$ 8.83)} &
        \makecell{5.93\\($\pm$ 4.14)} & \makecell{7.88\\($\pm$ 5.57)} &
        \makecell{\textbf{1.98}\\($\pm$ 1.28)} & \makecell{3.01\\($\pm$ 1.89)} &
        \makecell{\textbf{1.98}\\($\pm$ 1.29)} & \makecell{3.01\\($\pm$ 1.89)} &
        \makecell{2.06\\($\pm$ 1.58)} & \makecell{\textbf{2.99}\\($\pm$ 2.07)} \\ \cmidrule{2-11}
        Rotterdam &
        \makecell{57.95\\($\pm$ 40.14)} & \makecell{67.40\\($\pm$ 47.14)} &
        \makecell{22.36\\($\pm$ 18.14)} & \makecell{31.21\\($\pm$ 24.98)} &
        \makecell{8.57\\($\pm$ 6.31)} & \makecell{12.37\\($\pm$ 8.77)} &
        \makecell{8.58\\($\pm$ 6.35)} & \makecell{12.39\\($\pm$ 8.82)} &
        \makecell{\textbf{8.13}\\($\pm$ 4.77)} & \makecell{\textbf{11.72}\\($\pm$ 6.59)} \\ \bottomrule
     \end{tabular}
     }
    }
    \label{tab:perf-benchmark}
\end{table*}

To assess the impact of modeling granularity, we evaluated our r‑MoDE-Boost with a set of district‑specific XGBoost models, an approach proposed in \cite{TziorvasSharedMob}. In the local configuration, each model is trained exclusively on the time‑series of a single spatial entity, thereby eliminating cross‑district noise but also discarding the inter‑district dependencies that arise from spill‑over demand and correlated traffic flows. Conversely, MoDE-Boost leverages a unified feature‑engineering pipeline and a larger pooled training set, enabling it to capture commuting patterns and other spatial correlations that are inaccessible to the local models. 

\begin{table*}[h!]
    \fontsize{14}{8}\selectfont
    \centering
    \caption{Performance comparison of the r-MoDE-Boost, local XGBoost models, and TimeGPT for the 60 min. forecasting horizon across Rotterdam districts (the lowest error in bold type).}
    \renewcommand{\arraystretch}{1.5}
    \scriptsize
     \resizebox{\linewidth}{!}{%
        \begin{tabular}{lcccccccc}
            \toprule
            \multirow{2}{*}{\textbf{District}} & 
            \multirow{2}{*}{\textbf{Avg.\ Pop.}} &
            \multicolumn{2}{c}{\textbf{r‑MoDE‑Boost}} &
            \multicolumn{2}{c}{\textbf{Local XGBoost}} &
            \multicolumn{2}{c}{\textbf{TimeGPT}} \\ 
            \cmidrule(lr){3-4} \cmidrule(lr){5-6} \cmidrule(lr){7-8}
            & & \textbf{MAE} & \textbf{RMSE} &
            \textbf{MAE} & \textbf{RMSE} &
            \textbf{MAE} & \textbf{RMSE} \\
            \midrule
            Bedrijvenpark Noord‑West & 16  & \textbf{0.69} & \textbf{1.13} & 1.18 & 1.48 & 0.95 & 1.59 \\
            Charlois                 & 248 & 8.64  & 13.29 & \textbf{7.76} & \textbf{11.56} & 12.52 & 17.11 \\
            Delfshaven               & 398 & \textbf{9.73} & \textbf{13.01} & 11.14 & 14.75 & 22.15 & 26.51  \\
            Feijenoord               & 385 & \textbf{8.43} & \textbf{11.23} & 8.67 & 11.30 & 14.53 & 18.01  \\
            Hillegersberg‑Schiebroek & 435 & 14.08 & 21.50 & \textbf{12.03} & \textbf{18.22} & 17.06 & 23.41  \\
            Kralingen‑Crooswijk      & 530 & \textbf{12.03} & \textbf{17.29} & 12.37 & \textbf{16.99} & 17.08 & 23.89  \\
            Nieuw Mathenesse         & 39  & 2.48 & 4.54  & \textbf{2.43}  & \textbf{3.61} & 3.01  & 4.90   \\
            Noord                    & 338 & \textbf{8.87} & \textbf{12.53} & 9.15  & \textbf{12.47} & 15.35 & 19.43  \\
            Overschie                & 39  & 8.33  & 12.13 & 3.29 & 4.53 & \textbf{3.05}  & \textbf{3.96}   \\
            Prins Alexander          & 319 & 6.84  & 9.56  & \textbf{6.66} & \textbf{9.08} & 9.67  & 12.69  \\
            Rotterdam Centrum        & 621 & \textbf{17.41} & \textbf{23.29} & 20.02 & 26.68 & 34.37 & 44.74  \\
            Spaanse Polder           & 33  & \textbf{1.67} & \textbf{2.47} & 2.12 & 2.77 & 2.43  & 3.50   \\
            \^{I}Jsselmonde          & 184 & \textbf{6.51} & \textbf{10.45} & 40.33 & 69.06 & 7.06  & 10.18  \\ \midrule
            \textbf{Average $\mu \pm \sigma$} & & 
            \textbf{\makecell[c]{8.13 \\ ± 4.77}} &
            \textbf{\makecell[c]{11.56 \\ ± 6.53}} &
            \makecell[c]{10.55 \\ ± 10.36} & 
            \makecell[c]{15.58 \\ ± 17.58} &
            \makecell[c]{12.25 \\ ± 9.47} &
            \makecell[c]{16.15 \\ ± 12.07} \\
            \bottomrule
        \end{tabular}
     }
    \label{tab:performance-comparison}
\end{table*}

Across the thirteen Rotterdam districts, MoDE-Boost consistently attains the lowest MAE in the majority of cases, especially in districts with moderate to high average populations where the pooled training data furnish a richer representation of temporal dynamics. In these densely populated areas our approach also delivers comparable—or modestly superior RMSE, indicating greater robustness to occasional outliers. Conversely, the locally trained models achieve marginally lower MAE (and occasionally lower RMSE) in a limited subset of districts that are either sparsely populated or exhibit idiosyncratic demand patterns that are not well captured by the aggregated feature set. In such low‑data districts the exclusive focus on a single district’s time‑series enables the model to fine‑tune to local nuances (e.g., micro‑scale land‑use changes or isolated events), thereby offsetting the advantage of the global correlation structure. Nonetheless, the performance gain of the local models is relatively modest when weighed against their substantially larger footprint—approximately $11\times$ the size of MoDE-Boost—and the additional hyper‑parameter tuning required for each independent model, which together lead to considerably longer training times. These findings suggest that the our approach offers a more favourable trade‑off between predictive accuracy and computational efficiency for city‑wide deployment, while a hybrid strategy that selectively employs local models for a few outlier districts may be justified if marginal accuracy improvements are critical.

\subsubsection{Assessing the point-based variant of the MoDE-Boost framework}\label{sec:pSMDF}
In the second part of our experimental study, we compare our approach with a state-of-the-art model \cite{FengLiuASTN}. For the evaluation purposes, we use the "Point" group of datasets described in Table \ref{tab:data-desc}.

Our empirical evaluations on the CitiNYC and Divvy datasets show that the proposed r‑MoDE-Boost achieves performance that is essentially on par with current state‑of‑the‑art models such as the Adaptive Spatial‑Temporal Network (ASTN). To ensure a fair comparison, we reproduced the experimental protocol described in \cite{FengLiuASTN}: the full trip‑history (January 1, 2021 – June 1, 2022) was split into training, validation, and test sets using a 70:20:10 ratio, and the identical preprocessing pipeline (described in Section \ref{sec:data-desc}) was applied. Forecasts were generated for a 60‑minute horizon (i.e., one hour ahead) and evaluated using MAE and RMSE. As illustrated in Table \ref{tab:gbdp_vs_astn}, r‑MoDE-Boost yields a modest but consistent advantage over ASTN across both metrics in both datasets, indicating that our proposed model yields more robust predictions.

\begin{table*}[h]
    \centering
    \caption{Performance comparison of r-MoDE-Boost and ASTN (as reported in \cite{FengLiuASTN}) models on CitiNYC and Divvy datasets, measured by MAE and RMSE. Note: in this experiment, the error scores are reported in standardized units (demand counts).}
    \label{tab:gbdp_vs_astn}
    \renewcommand{\arraystretch}{1.4}
    \begin{tabular}{@{}lcccccccc@{}}
        \toprule
        \multirow{2}{*}{Dataset} & \multicolumn{2}{c}{ASTN} & \multicolumn{2}{c}{MoDE-Boost} \\ \cmidrule(l){2-3} \cmidrule(l){4-5}
                & MAE   & RMSE  & MAE     & RMSE \\ \midrule
        CitiNYC    & 1.20 & 2.22 &\textbf{1.13} & \textbf{1.78} \\
        Divvy      & 0.77 & 1.43 &\textbf{0.76} & \textbf{1.25} \\

        \bottomrule
    \end{tabular}
\end{table*}

\subsection{Experiments at the Edge}\label{sec:edgeexperiments}

The proposed MoDE-Boost model is not only accurate and robust as shown in Section \ref{sec:exp-results}, but also exceptionally well-suited for deployment at the edge, particularly on resource-constrained devices, such as the Raspberry Pi 5 with 16 GB of RAM. This section presents the results of edge-based inference experiments conducted on this platform, emphasizing the model's ultra-low inference latency and minimal memory footprint as key enablers for real-time, on-device micro-mobility demand forecasting.
The model was deployed in a realistic edge environment using lightweight Python inference scripts and optimized XGBoost runtime configurations. Inference times were measured across all prediction horizons (5, 15, 30, and 60 min.) and for both regression and classification variants, across districts in Amsterdam, Rotterdam, and The Hague.

\begin{figure}[h!]
    \centering
    \includegraphics[width=.8\columnwidth]{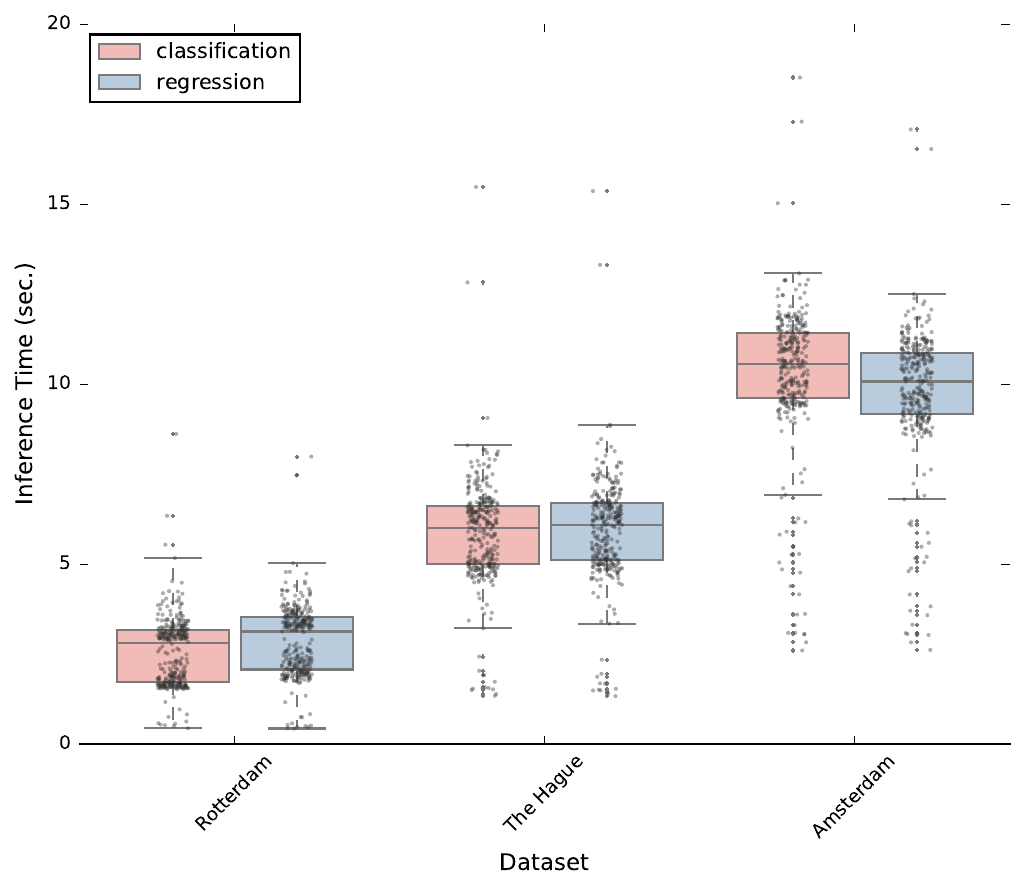}
    \caption{Inference time of the MoDE-Boost model (both variants) over three different datasets and a 60 min. forecasting horizon, running at an Edge environment.}
    \label{fig:edge_inference}
\end{figure}

The results, summarized in Figure~\ref{fig:edge_inference}, reveal the level of efficiency of our method. On average, our model processes and infers forecasting of approximately 3500 records per batch, at city level in under 5-15 seconds, depending on the city of interest, which makes an inference time of 1.5 to 4.5 msec per record. If we consider that the authority that collects this data is using a 60-second sampling time window (see Section \ref{sec:setup}), we can infer that the reported performance enables near-instantaneous decision-making across multiple districts, making the model ideal for real-time applications such as dynamic vehicle redistribution, congestion alerts, or adaptive traffic signal control.

Crucially, the memory usage remains consistently low across all components. The peak memory consumption per individual component (i.e., per district-model pair) stays below 300 MB, with the total memory footprint for both models (classification and regression) residing at an edge node staying under 500 MB, well within the 16 GB available on the Raspberry Pi 5. These results underline the need for approaches like the one proposed in this paper that utilize considerably fewer resources than deep learning models, which often require several GB of RAM even for inference, and may fail on such edge hardware altogether.

Overall, our gradient-boosted approach achieves true edge readiness: high accuracy, ultra-low latency, and minimal resource consumption, making it uniquely suitable for deployment in real-world urban environments where computational resources are limited. As demonstrated earlier, a fleet manager could run the full forecasting pipeline on a local Raspberry Pi at a depot, enabling real-time, privacy-preserving predictions without relying on cloud connectivity. This reduces latency, enhances data sovereignty and privacy, and ensures resilience in case of network outages. Moreover, the low memory and computational demands allow for scalable deployment across multiple edge nodes—such as city district controllers, smart kiosks, or embedded traffic systems—without significant infrastructure overhead. The ability to run multiple district-specific models simultaneously on a single low-power device underscores the practical viability of our method in large-scale, distributed smart city infrastructures. 
\section{Conclusions}\label{sec:conclusions}

This study investigated spatio‑temporal forecasting techniques for predicting urban shared micromobility demand. 
The model adopts a broader perspective on the multiple timeseries demand forecasting problem. Instead of focusing solely on a spatial entity, it merges all timeseries from within the selected metropolitan area. These are organized into a single multi-indexed dataset, where the first index is associated with a corresponding district identifier and the second level of indices corresponds to the timestamps. This design enables the model to exploit patterns across regions and supports a single model trained jointly on the entire spatial domain.

By jointly modeling spatial dependencies and temporal dynamics, the proposed framework captures both localized demand patterns and inter‑district flows, thereby delivering accurate forecasts across multiple horizons. In summary, the principal contributions of this paper are twofold: (i) we proposed MoDE-Boost, a flexible feature‑engineering addressing the SMDF problem in two variants, as a regression or a classification task, and (ii) we demonstrated the deployment of efficient gradient‑boosting models that achieve high predictive performance while incurring modest computational cost, rendering the approach scalable to city‑wide applications. Through an extensive empirical evaluation on a variety of real-world datasets, we demonstrated that gradient‑boosted demand predictors attain competitive results for both regression and classification tasks at all horizons; although predictive accuracy declines slightly for longer lead times, the models remain robust, confirming their practical suitability for real‑world mobility‑forecasting scenarios.

Future work will extend the current methodology in several directions. Incorporating weather variables—following related approaches \cite{Lee2024, s22031060}—should provide valuable external context and potentially improve forecast precision. Moreover, we intend to broaden the classification scheme beyond three classes and also implement dynamic label‑thresholding strategies to mitigate the challenges posed by highly skewed demand distributions. These extensions will further enhance the adaptability and reliability of the forecasting system for heterogeneous urban environments.

\bmhead{Acknowledgments}
This work was supported in part by the Horizon Framework Programme of the European Union under grant agreement No. 101093051 (EMERALDS; https://www.emeralds-horizon.eu/).

\section{Code and Data availability}
Our source code, including details about the open datasets used in our experimental study, is publicly available for reproducibility purposes om GitHub\footnote{\url{https://github.com/DataStories-UniPi/Shared-Mobility.git}}.

\bibliography{sn-bibliography-raw}

@article{Wu2019ACS,
	title        = {{A Comprehensive Survey on Graph Neural Networks}},
	author       = {Zonghan Wu and Shirui Pan and others},
	year         = 2019,
	pages        = {4--24},
	journal      = {IEEE Transactions on Neural Networks and Learning Systems},
	doi          = {https://doi.org/10.1109/TNNLS.2020.2978386},
	volume       = 32,
}

@article{Caggiani2018AMF,
	title        = {{A modeling framework for the dynamic management of free-floating bike-sharing systems}},
	author       = {Leonardo Caggiani and Rosalia Camporeale and Michele Ottomanelli and others},
	year         = 2018,
	pages        = {159--182},
	journal      = {Transportation Research Part C: Emerging Technologies},
	doi          = {https://doi.org/10.1016/j.trc.2018.01.001},
	volume       = 87,
	issn         = {0968-090x},
}

@article{Hu2021ASL,
	title        = {{A spatio-temporal LSTM model to forecast across multiple temporal and spatial scales}},
	author       = {Fearghal O'Donncha and Yihao Hu and others},
	year         = 2022,
	pages        = 101687,
	journal      = {Ecological Informatics},
	doi          = {https://doi.org/10.1016/j.ecoinf.2022.101687},
	volume       = 69,
	issn         = {1574-9541},
}

@article{Yang2024ASO,
	title        = {{A Survey on Diffusion Models for Time Series and Spatio-Temporal Data}},
	author       = {Yiyuan Yang and Ming Jin and others},
	year         = 2024,
	journal      = {arXiv: 2404.18886},
	doi          = {10.48550/arXiv.2404.18886},
}

@inproceedings{Bergstra2011TPESampler,
	title        = {Algorithms for hyper-parameter optimization},
	author       = {Bergstra, James and Bardenet, R\'{e}mi and Bengio, Yoshua and others},
	booktitle    = {Proceedings of the 25th International Conference on Neural Information Processing Systems (NIPS)},
	year         = 2011,
	isbn         = 9781618395993,
	publisher    = {Curran Associates Inc.},
	pages        = {2546–2554},
	numpages     = 9,
}

@article{FengLiuASTN,
	title        = {{An Adaptive Spatial-Temporal Method Capturing for Short-Term Bike-Sharing Prediction}},
	author       = {Feng, Jiahui and Liu, Hefu},
	year         = 2024,
	pages        = {16761--16774},
	journal      = {IEEE Transactions on Intelligent Transportation Systems},
	doi          = {https://doi.org/10.1109/TITS.2024.3406682},
	volume       = 25,
	number       = 11,
}

@article{s22031060,
	title        = {{Bike-Sharing Demand Prediction at Community Level under COVID-19 Using Deep Learning}},
	author       = {Mehdizadeh Dastjerdi, Aliasghar and Morency, Catherine},
	year         = 2022,
	journal      = {Sensors},
	doi          = {https://doi.org/10.3390/s22031060},
	volume       = 22,
	number       = 3,
	article-number = 1060,
	issn         = {1424-8220},
}

@inproceedings{Lee2024,
	title        = {Bike-sharing Demand Prediction based on Artificial Intelligence Algorithm Using Weather Data},
	author       = {Lee, Yebeen and Son, Hyungju and Ahn, Jiin and others},
	booktitle    = {Proceedings of the IEEE International Conference on Consumer Electronics (ICCE)},
	year         = 2024,
	pages        = {1--6},
	doi          = {10.1109/ICCE59016.2024.10444462},
}

@article{Hooker2020Bias,
	title        = {{Characterising Bias in Compressed Models}},
	author       = {Sara Hooker and Nyalleng Moorosi and Gregory Clark and others},
	year         = 2020,
	journal      = {arXiv: 2010.03058},
	doi          = {https://arxiv.org/abs/2010.03058},
}

@article{Lee2019TGNet,
	title        = {{Demand Forecasting from Spatiotemporal Data with Graph Networks and Temporal-Guided Embedding}},
	author       = {Doyup Lee and Suehun Jung and Yeongjae Cheon and others},
	year         = 2019,
	journal      = {arXiv: 1905.10709},
	doi          = {https://arxiv.org/abs/1905.10709},
}

@article{Li2017DiffusionCR,
	title        = {{Diffusion Convolutional Recurrent Neural Network: Data-Driven Traffic Forecasting}},
	author       = {Yaguang Li and Rose Yu and others},
	year         = 2017,
	journal      = {arXiv: 1707.01926},
	doi          = {10.48550/arXiv.1707.01926},
}

@article{Sohrabi2021DynamicBS,
	title        = {{Dynamic bike sharing traffic prediction using spatiotemporal pattern detection}},
	author       = {Soheil Sohrabi and Alireza Ermagun},
	year         = 2021,
	pages        = 102647,
	journal      = {Transportation Research Part D: Transport and Environment},
	doi          = {https://doi.org/10.1016/j.trd.2020.102647},
	volume       = 90,
	issn         = {1361-9209},
}

@article{Vemuri2024EnhancingPT,
	title        = {{Enhancing Public Transit System Through AI and IoT}},
	author       = {Naveen Vemuri and Venkata Manoj Tatikonda and Naresh Thaneeru},
	year         = 2024,
	pages        = {1057–1071},
	journal      = {International Journal of Scientific Research and Management (IJSRM)},
	doi          = {https://doi.org/https://doi.org/10.18535/ijsrm/v12i02.ec07},
	volume       = 12,
	number       = {02},
	month        = {Feb.},
}

@article{Croston1972,
	title        = {{Forecasting and stock control for intermittent demand}},
	author       = {Croston, J.D.},
	year         = 1972,
	pages        = {289 – 303},
	journal      = {Operational Research Quarterly},
	doi          = {10.1057/jors.1972.50},
	volume       = 23,
	number       = 3,
}

@article{HoltWintersSES,
	title        = {{Forecasting Sales by Exponentially Weighted Moving Averages}},
	author       = {Winters, Peter R.},
	year         = 1960,
	publisher    = {INFORMS},
	pages        = {324–342},
	numpages     = 19,
	journal      = {Manage. Sci.},
	doi          = {10.1287/mnsc.6.3.324},
	volume       = 6,
	number       = 3,
	issn         = {0025-1909},
	url          = {https://doi.org/10.1287/mnsc.6.3.324},
	month        = apr,
}

@article{Zheng2019GMANAG,
	title        = {{GMAN: A Graph Multi-Attention Network for Traffic Prediction}},
	author       = {Zheng, Chuanpan and Fan, Xiaoliang and others},
	year         = 2020,
	pages        = {1234--1241},
	journal      = {Proceedings of the AAAI Conference on Artificial Intelligence},
	doi          = {https://doi.org/10.1609/aaai.v34i01.5477},
	month        = {Apr.},
	volume       = 34,
	number       = {01},
}

@article{GNNPTDP2020,
	title        = {{Graph Neural Network for Robust Public Transit Demand Prediction}},
	author       = {Li, Can and Bai, Lei and Liu, Wei and others},
	year         = 2022,
	pages        = {4086--4098},
	journal      = {IEEE Transactions on Intelligent Transportation Systems},
	doi          = {https://doi.org/10.1109/TITS.2020.3041234},
	volume       = 23,
	number       = 5,
}

@incollection{MLADENOVIC202112,
	title        = {{Mobility as a Service}},
	author       = {Milo\v{s} N. Mladenovi\'{c}},
	booktitle    = "International Encyclopedia of Transportation",
	year         = 2021,
	isbn         = "978-0-08-102671-7",
	publisher    = "Academic Press",
	month        = may,
	language     = "English",
	editor       = "Roger Vickerman",
}

@book{Pelekis2014-fw,
	title        = {{Mobility Data Management and Exploration}},
	author       = {Pelekis, Nikos and Theodoridis, Yannis},
	year         = 2014,
	publisher    = {Springer},
	doi          = {https://doi.org/10.1007/978-1-4939-0392-4},
	month        = mar,
	language     = {en},
}

@inproceedings{optuna2019,
	title        = {Optuna: A Next-generation Hyperparameter Optimization Framework},
	author       = {Akiba, Takuya and Sano, Shotaro and Yanase, Toshihiko and others},
	booktitle    = {Proceedings of the 25th ACM SIGKDD International Conference on Knowledge Discovery \& Data Mining (KDD)},
	year         = 2019,
	isbn         = 9781450362016,
	pages        = {2623–2631},
	numpages     = 9,
	doi          = {https://doi.org/10.1145/3292500.3330701},
	location     = {Anchorage, AK, USA},
}

@article{ZhangSTResNet,
	title        = {{Predicting citywide crowd flows using deep spatio-temporal residual networks}},
	author       = {Junbo Zhang and Yu Zheng and Dekang Qi and others},
	year         = 2018,
	pages        = {147--166},
	journal      = {Artificial Intelligence},
	doi          = {https://doi.org/10.1016/j.artint.2018.03.002},
	volume       = 259,
	issn         = {0004-3702},
}

@inproceedings{YangLiProphetBiLSTM,
	title        = {{Prediction method of shared bicycle traffic based on Prophet-BiLSTM combined model}},
	author       = {Yang, Gang and Li, Haiming},
	booktitle    = {{Proceedings of the 10th International Conference on Information Technology: IoT and Smart City,(ICIT)}},
	year         = 2023,
	isbn         = 9781450397438,
	pages        = {251–256},
	numpages     = 6,
	doi          = {https://doi.org/10.1145/3582197.3582239},
	location     = {Shanghai, China},
}

@article{Yao2019STDN,
	title        = {{Revisiting Spatial-Temporal Similarity: A Deep Learning Framework for Traffic Prediction}},
	author       = {Yao, Huaxiu and Tang, Xianfeng and Wei, Hua and others},
	year         = 2019,
	pages        = {5668--5675},
	journal      = {Proceedings of the AAAI Conference on Artificial Intelligence},
	doi          = {https://doi.org/10.1609/aaai.v33i01.33015668},
	volume       = 33,
	month        = {Jul.},
}

@proceedings{TziorvasSharedMob,
	title        = {{Shared Micro-mobility Demand Forecasting using Gradient Boosting methods}},
	author       = {Tziorvas, Antonios and Theodoropoulos, George S. and Theodoridis, Yannis},
	year         = 2025,
	series       = {{7th International Workshop on Big Mobility Data Analytics (BMDA)}},
	volume       = 3946,
	url          = {https://ceur-ws.org/Vol-3946/BMDA-4.pdf},
}

@article{Li2022STMN,
	title        = {{Short-Term Forecast of Bicycle Usage in Bike Sharing Systems: A Spatial-Temporal Memory Network}},
	author       = {Li, Xinyu and Xu, Yang and Chen, Qi and others},
	year         = 2022,
	pages        = {10923--10934},
	journal      = {IEEE Transactions on Intelligent Transportation Systems},
	doi          = {https://doi.org/10.1109/TITS.2021.3097240},
	volume       = 23,
	number       = 8,
}

@inproceedings{STClusterFFBSS,
	title        = {{Spatio-temporal Clustering and Forecasting Method for Free-Floating Bike Sharing Systems}},
	author       = {Caggiani, Leonardo and Ottomanelli, Michele and Camporeale, Rosalia and others},
	booktitle    = {Advances in Systems Science},
	year         = 2017,
	isbn         = {978-3-319-48944-5},
	publisher    = "Springer International Publishing",
	pages        = {244--254},
	doi          = {https://doi.org/10.1007/978-3-319-48944-5_23},
}

@inproceedings{Yu2017SpatiotemporalGC,
	title        = {{Spatio-Temporal Graph Convolutional Networks: A Deep Learning Framework for Traffic Forecasting}},
	author       = {Yu, Bing and Yin, Haoteng and others},
	booktitle    = {Proceedings of the 27th International Joint Conference on Artificial Intelligence},
	year         = 2018,
	pages        = {3634–3640},
	doi          = {https://doi.org/10.24963/ijcai.2018/505},
	month        = jul,
	collection   = {Ijcai-2018},
}

@article{Geng2019STMGCN,
	title        = {{Spatiotemporal Multi-Graph Convolution Network for Ride-Hailing Demand Forecasting}},
	author       = {Geng, Xu and Li, Yaguang and others},
	year         = 2019,
	pages        = {3656--3663},
	journal      = {Proceedings of the AAAI Conference on Artificial Intelligence},
	doi          = {https://doi.org/10.1609/aaai.v33i01.33013656},
	month        = {Jul.},
	volume       = 33,
	number       = {01},
}

@article{Liao2022TaxiDF,
	title        = {{Taxi demand forecasting based on the temporal multimodal information fusion graph neural network}},
	author       = {Wenxiong Liao and Bi Zeng and others},
	year         = 2022,
	pages        = {12077--12090},
	journal      = {Applied Intelligence},
	doi          = {https://doi.org/10.1007/s10489-021-03128-1},
	volume       = 52,
}

@article{garza2023timegpt1,
	title        = {TimeGPT-1},
	author       = {Azul Garza and Cristian Challu and Max Mergenthaler-Canseco},
	year         = 2024,
	journal      = {arXiv: 2310.03589},
	doi          = {https://arxiv.org/abs/2310.03589},
}

@article{LibanoQuant2020,
	title        = {{Understanding the Impact of Quantization, Accuracy, and Radiation on the Reliability of Convolutional Neural Networks on FPGAs}},
	author       = {Libano, F. and Wilson, B. and Wirthlin, M. and others},
	year         = 2020,
	pages        = {1478--1484},
	journal      = {IEEE Transactions on Nuclear Science},
	doi          = {https://doi.org/10.1109/TNS.2020.2983662},
	volume       = 67,
	number       = 7,
}

@inproceedings{Grinsztajn22,
	title        = {{Why do tree-based models still outperform deep learning on typical tabular data?}},
	author       = {Grinsztajn, L\'{e}o and Oyallon, Edouard and others},
	booktitle    = {Proceedings of the 36th International Conference on Neural Information Processing Systems},
	year         = 2022,
	isbn         = 9781713871088,
	pages        = {507--520},
	numpages     = 14,
	doi          = {https://doi.org/10.48550/arXiv.2207.08815},
	series       = {(NIPS)},
	articleno    = 37,
}

\end{document}